# Virtual Urbanism: An AI-Driven Framework for Quantifying Urban Identity.
# A Tokyo-Based Pilot Study Using Diffusion-Generated Synthetic Environments

Glinskaya Maria[1]

[1]Department of Architecture, The University of Tokyo, Tokyo, Japan

**Abstract**

This paper introduces Virtual Urbanism (VU), a multimodal AI-driven analytical framework for quantifying urban identity through the medium of synthetic urban replicas. The framework aims to advance computationally tractable urban identity metrics. To demonstrate feasibility, the pilot study Virtual Urbanism and Tokyo Microcosms is presented. A pipeline integrating Stable Diffusion and LoRA models was used to produce synthetic replicas of nine Tokyo areas rendered as dynamic synthetic urban sequences, excluding existing orientation markers to elicit core identity-forming elements. Human-evaluation experiments (I) assessed perceptual legitimacy of replicas; (II) quantified area-level identity; (III) derived core identity-forming elements. Results showed a mean identification accuracy of ~81%, confirming the validity of the replicas. Urban Identity Level (UIL) metric enabled assessment of identity levels across areas, while semantic analysis revealed culturally embedded typologies as core identity-forming elements, positioning VU as a viable framework for AI-augmented urban analysis, outlining a path toward automated, multi-parameter identity metrics.

**Keywords**

generative artificial intelligence, latent diffusion model, low-rank adaptation model, urban perception, urban identity

## 1. Introduction

Recent advances in Artificial Intelligence (AI) have rapidly expanded its presence across diverse domains, ranging from natural-language processing supported by Large Language Models (LLMs) to visual synthesis enabled by Diffusion Models (DMs). These developments are reshaping creative, analytical, and interpretive methodologies across architectural and urban studies [1], allowing researchers to move beyond conventional—often reductionist, resource-bound—modes of urban study, such as continuous elevation drawing [2], toward new, data-driven methods that focus on semantic understanding and AI-assisted decoding of urban identity [3].

In parallel, contemporary urban development is increasingly shaped by forces of globalization, standardization, and economic optimization, where project priorities often centre on production efficiency, market positioning, and replicable design strategies [4]. Within such conditions, locally embedded *urban identity*—a naturally evolved configuration of perceptual cues and locally distinctive features, shaped by historical layering and cultural narratives, that define an area's visual–spatial coherence and perceived authenticity—tend to be treated as secondary considerations, despite their critical role in shaping lived experience and cultural continuity. This has heightened the importance of developing computational methods capable of documenting, analysing, and preserving urban identity before it becomes diluted or lost. As a result, the adaptation of machine-learning techniques to urban-perception research has become an emerging trajectory.

While a number of studies have begun to incorporate AI-generated content as part of their analytical framework [5],[6], a recurring omission emerges: the unique analytical affordance offered by generative AI—its ability to generate large volumes of controlled, high-dimensional variation—is rarely leveraged as an analytical asset. Existing approaches still mainly rely on a small number of manually selected outputs drawn from what is, in principle, a theoretically unbounded generative space. A more substantial methodological gap concerns the fact that AI is still predominantly used to supplement conventional analytical workflows rather than to establish new ones, leaving its potential to function as a multimodal, variation-rich framework for urban analysis largely unexplored.

This work addresses these limitations by positioning generative AI as the methodological core of a novel analytical framework, shifting urban inquiry from static imagery to dynamic synthetic environments and from AI-assisted to AI-driven analysis. In doing so, it enables the systematic revelation of previously inaccessible patterns in urban identity, with potential applicability across analytical urban research, applied design practice, computational urban identity archiving, and the identification of identity-rich urban environments.

The primary visual analytical materials and fine-tuned generative models produced in this study are publicly available; details are provided in the Data Availability section.

## 1.1 Core Methodological Approach

The main contribution of this paper is the introduction and demonstration of *Virtual Urbanism (VU)*, a multimodal AI-driven analytical framework for quantifying urban identity through the medium of synthetic urban replicas. *Synthetic urban replicas* are defined here as generatively produced, non-metric, non-literal reconstructions of real urban environments—approximate representational models deliberately structured to function as an analytical medium, created to enable forms of insight not directly attainable through the real environment itself.

The main analytical aspect of synthetic replicas is their *dynamic quality*—here defined as a state of constant fluctuation in the visual representation of form, enabled by AI-driven synthetic generative variation. In this work's pilot study, this dynamic quality was operationalized by presenting the replicas as dynamic synthetic urban sequences. A *dynamic synthetic urban sequence* is defined as a series of images rendered within the synthetic urban replica and subsequently re-generated through the diffusion model—each frame undergoing image-to-image generative transformation under a constrained, fine-tuned generative model—to introduce maximal AI-driven variation in the representation of the same form or environment. These frames are then assembled into a video sequence that produces dense perceptual fluctuation, resulting in a highly intensive dynamic analytical representational medium.

Accordingly, here the sequence of appearances becomes the primary analytical tool for identifying *core urban identity–forming elements*—recurrent perceptual cues, visual and spatial features, and typological references that function as primary anchors of an area's perceived authenticity and cohesive local urban identity—building on the concept of the *interrepresentational approach* originally introduced by Marcos Novak. This concept can be traced to Novak's essay *Liquid Architecture* (1991) [7], in which he reinterprets Sartre's *Principle of the Series* [8]—a philosophical position proposing that essence emerges through a sequence of appearances governed by internal logic rather than subjective perception. Under this interpretation, identity is not fixed or static, but revealed through *successive manifestations*.

Extending this conceptual line, the VU framework leverages generative AI to construct synthetic urban replicas of existing environments endowed with dynamic quality. By re-envisioning the urban environment as a sequence of successive manifestations generated through controlled, AI-driven variation of the same underlying form, VU aims to render the internal logic of urban identity perceptible through large volumes of systematic, high-dimensional repetition. In this formulation, AI-enabled successive generative appearances operate as an analytical mechanism for revealing core urban identity structures, effectively rearticulating Novak's original interrepresentational approach within contemporary urban research through the VU framework. The present pilot study operationalizes this conceptual logic in a preliminary form.

## 1.2 Research Overview

At the long-term level, the VU framework sets goals to advance the development of computationally tractable identity metrics—primarily encompassing (I) the formalization of urban identity as a measurable and cross-comparable metric, and (II) the quantification of the elements that define urban identity and the determination of their relative weight.

This paper represents the first applied stage of that agenda, specifically investigating the following objectives:
(a) the capacity of generative models in replicating real urban settings;
(b) the perceptual validity of these replications through human perceptual evaluation experiments;
(c) the applicability of AI-driven synthetic urban replicas as an analytical medium for studying urban identity at the visual, streetscape level.

The remainder of the paper is structured as follows. Section 2, Related Works, reviews prior research. Section 3, Pilot Study "Virtual Urbanism and Tokyo Microcosms", introduces the pilot study through which the VU framework is demonstrated. Where Section 4, Experiment, describes the method for constructing synthetic urban replicas, corresponding to objective (a). Section 5, Evaluation, outlines the method for extracting analytical data through human perceptual evaluation addressing objective (b). And Section 6, Results and Discussion, analyses the acquired data to generate insights addressing objective (c). Section 7, Limitations and Challenges, reflects on the methodological and technical constraints identified. Section 8: Conclusion, summarizes the overall outcomes of the pilot study, and outlines directions for future development of the VU framework.

## 2. Related Works

This review contextualizes generative models in urban studies by surveying generative urban design (GUD), emerging AI-based urban perception research, and related applications in architectural heritage and local-identity preservation. Each domain reveals distinct methodological gaps, which collectively delineate the positioning of the present study within the broader field.

## 2.1 Generative Urban Design

Generative AI is currently finding growing implementation within architectural discourse, ranging from automated ideation in GUD [9] to simulation [10], visualization [11], and human-in-the-loop design evaluation [12]. Particularly widely used in practical implementation are vision-based generative models, which have recently experienced a shift from earlier versions—Generative Adversarial Networks (GANs) [13], operating through a generator–discriminator system—to more refined, controllable, and accessible Denoising Diffusion Models (DDMs), such as Latent Diffusion [14], featuring true text-to-image generation and accompanied by major simplifications in user interface (UI) design. This transition has catalyzed a decisive shift, transitioning the field from niche science to mainstream application.

While mostly utilized during the Schematic Design stage of architectural project development [15], applications of DMs can be found across a wide range of domains in both urban research and production, ranging from early-stage urban analysis [16] to controllable floorplan generation [17], architectural façade synthesis [18], visualization and rendering assistance [19], and further expanding into three-dimensional spatial generation [20].

Yet, as Hazbei and Cucuzzella [21] note, computational approaches often risk reducing architectural context to parametric constraints, overlooking the more intangible social, cultural, and perceptual dimensions that constitute the essence of place. Their observation underscores a critical gap: current GUD practices widely rely on AI for measurable optimization but often overlook the perceptual factors that define human experience of place. This work responds to the challenge outlined above by contributing to the empirical exploration of the capacity of generative models to articulate contextual qualities of the built environment and gauge urban identity, thereby enabling qualitative, higher-dimensional perceptual factors to be incorporated into measurable parameters that can, in turn, be reintegrated into future GUD workflows.

## 2.2 Urban Perception Studies

A parallel trajectory of AI integration has emerged within urban perception studies, where machine learning is employed to evaluate how humans perceive and emotionally respond to built environments. Representative studies include Zhang et al. [22], predicting perceptual attributes from street-view imagery; Liu et al. [23], employing geo-tagged photos to computationally reassess Lynch's imageability elements; and Ordonez and Berg [24], demonstrating cross-city prediction of perceptual judgments from visual data correlated with socioeconomic indicators.

What earlier generations of urban theorists pursued as qualitative exploration of city atmospheres—for instance, the once-unfinished ambitions of the Situationist International's psychogeographic maps and urban dérives [25] —is now, being reawakened, reinterpreted, and operationalized, translating affective and perceptual qualities of urban space into quantifiable visual data [22] and allowing perceptual urban phenomena to be examined computationally at metropolitan and city-wide scales [26].

Despite the rapid integration of AI-driven methods into this research trajectory, only limited studies have examined qualitative urban characteristics through artificially generated media and their alignment with human perception, such as Tanigawa and Kobayashi's analysis of AI-generated streetscape imagery and perceived authenticity [5], and Law et al.'s perceptual alignment studies of synthetic architectural façades [27]. Another tendency can be observed: with the rapid involvement of machine-based methods encouraging an accelerated expansion of scale, research is progressively zooming out from the ground-level urban resolution toward meta-scale perspectives, a pattern evident across the majority of studies reviewed above.

In response, this study takes a step back to explore meso-scale, context-sensitive applications of generative AI—operationalized specifically through the introduction of an AI-driven analytical medium—focusing on confined urban areas, thus counterbalancing the field's prevailing shift toward city-scale analysis.

## 2.3 Image-Generative AI for Architectural Heritage and Local Identity
### 2.3.1 GAN-based

The domain that can be positioned at the convergence of GUD and Urban Perception Studies is the emerging use of image-generative AI for preserving locality-specific architectural and cultural features. Early explorations in this area primarily relied on Generative Adversarial Networks (GANs). For instance, Bachl and Ferreira [28] proposed City-GAN, a conditional GAN trained to learn and reproduce architectural styles from city-specific image datasets. Steinfeld [6] introduced GAN-Loci, designed to extract and replicate the implicit spatial characteristics of urban areas, implementing StyleGAN model [29]. Ali and Lee [30] presented iFACADE—a CycleGAN-based [31] generator for urban infill, producing façades style-mixed from adjacent buildings. While, Sun et al. [32] proposed the CycleGAN-based method for identifying and reproducing historic architectural styles to support urban renovation. While GAN-based studies established important early groundwork for locality-aware generative workflows, their broader adoption has been limited by well-documented challenges, including training instability, lower detail fidelity, and restricted controllability [33].

### 2.3.2 Diffusion-based

DMs emerged as a more stable alternative, offering higher-resolution synthesis, multimodal conditioning, and greater flexibility. Within this trajectory, Latent Diffusion Models (LDMs) [14] have become the basis for research-oriented workflows, with open architecture supporting detailed conditioning, parameter control, and domain-specific fine-tuning. For instance, Law et al. [27] used Stable Diffusion (SD) to generate geographically plausible counterfactual façades and evaluated them against geographical, objective, and affective descriptors via AI- and human-based perceptual alignment. Jo et al. [19] developed an approach for generating alternative architectural design options incorporating regional stylistic characteristics using SD fine-tuned via Low-Rank Adaptation (LoRA) technique [34] enabling parameter-efficient domain specialization.

Several studies further demonstrate the suitability of SD and LoRA-based fine-tuning for heritage-oriented applications. Kuang et al. [35] introduced a framework integrating SD, LoRA, and ControlNet [36] to support automated architectural heritage preservation in urban renewal projects. Shin [37] employed diffusion-based workflows for heritage artefact 3D reconstruction. Guo et al. [38] applied LoRA-fine-tuned DM to the generative reconstruction of Yi ethnic embroidery patterns, illustrating the broader cultural-heritage relevance.

There is a simple but fundamental methodological distinction between the present study and the predominant body of work outlined above. In prior studies, generative models are used primarily for reproduction, with generative output treated as the end goal of the process. In contrast, this study reproduces in order to analyse. Synthetic replication is not the end goal but the means to investigate contextual cues and to uncover new perceptual insights into urban identity.

## 3. Pilot Study "Virtual Urbanism and Tokyo Microcosms"

To demonstrate the proposed framework's potential, a pilot implementation—*Virtual Urbanism and Tokyo Microcosms*—is presented, operationalized across nine Tokyo urban areas. In this study, the term *Microcosms* was introduced to avoid administrative framings such as "districts" or "cities," and instead denote perceptually coherent urban units. Each study area was defined at a neighborhood-scale spatial extent, with boundaries approximating a walkable radius of approximately 1.5 km.

Tokyo was selected as the pilot site due to its intrinsic spatial complexity, defined by high visual density, a multilayered urban fabric, and rapidly changing contextual conditions. Identifying core urban identity–forming elements here presents a significant analytical challenge, making the city an appropriate testbed for evaluating the capacities of the VU framework.

The selection of the case-study areas was intended to capture a spectrum of urban development trajectories, ranging from organically evolved, community-driven environments to areas shaped by centralized planning and corporate redevelopment. The nine case-study areas thus comprised: Shimokitazawa, Harajuku, Yanesen, Kagurazaka, Asakusa, Ueno, Shibuya, Ikebukuro, and Roppongi.

### 3.1 Pilot Study Workflow and Objectives

The pilot study consisted of three sequential components, corresponding to Sections 4, Experiment; 5, Evaluation; and 6, Results and Discussion. Together, these sections outline the construction, assessment, and implementation of AI-driven synthetic urban replicas as an analytical medium within the VU framework. (Figure 1)

Section 4, Experiment, details the methodology for constructing a Tokyo-based synthetic urban replica as a single three-dimensional base, subsequently operationalized into nine dynamic synthetic urban sequences as the study's primary analytical medium. The analytical advantage of this medium over direct real-world observation is twofold. First, its primary feature is dynamic quality, defined earlier in alignment with Novak's interrepresentational approach. Dynamic repetition functioning as a perceptual amplifier, enabling latent patterns to emerge through sustained exposure rather than static inspection. Second, both the replica and the corresponding sequences are deliberately constructed to omit existing orientation markers. *Orientation markers*—here defined as visually salient reference elements such as landmarks, signage, or other distinctive features that facilitate immediate place identification—are removed to eliminate recognition driven by prior familiarity, collectively foregrounding deeper contextual cues.

The resulting sequences thus operate as a controlled, landmark-absent analytical medium through which urban identity can be examined via human perceptual evaluation, the methodological procedure of which is detailed in Section 5, Evaluation. Section 6, Results and Discussion, subsequently presents and interprets the collected data.

### Alignment with Research Objectives

Within this structure, the pilot study directly addresses previously defined research objectives (a) and (b) by testing both the technical capacity and the perceptual validity of AI-driven synthetic urban replicas. It further advances objective (c)—the assessment of AI-driven synthetic urban replicas as an analytical medium—through two focused sub-objectives, designed specifically to implement the previously outlined long-term goals of the VU framework:

(cI) to evaluate and compare Urban Identity Levels (UIL) across the selected study areas—operationalized through the use of identification Accuracy Rate parameter per sequence as preliminary metric;

(cII) to define the core urban identity–forming elements of each area—derived through semantic analysis of cue-emergent free-text survey responses collected during human perceptual evaluation sessions.

Their implementation and resulting analyses are discussed in Sections 5 and 6.

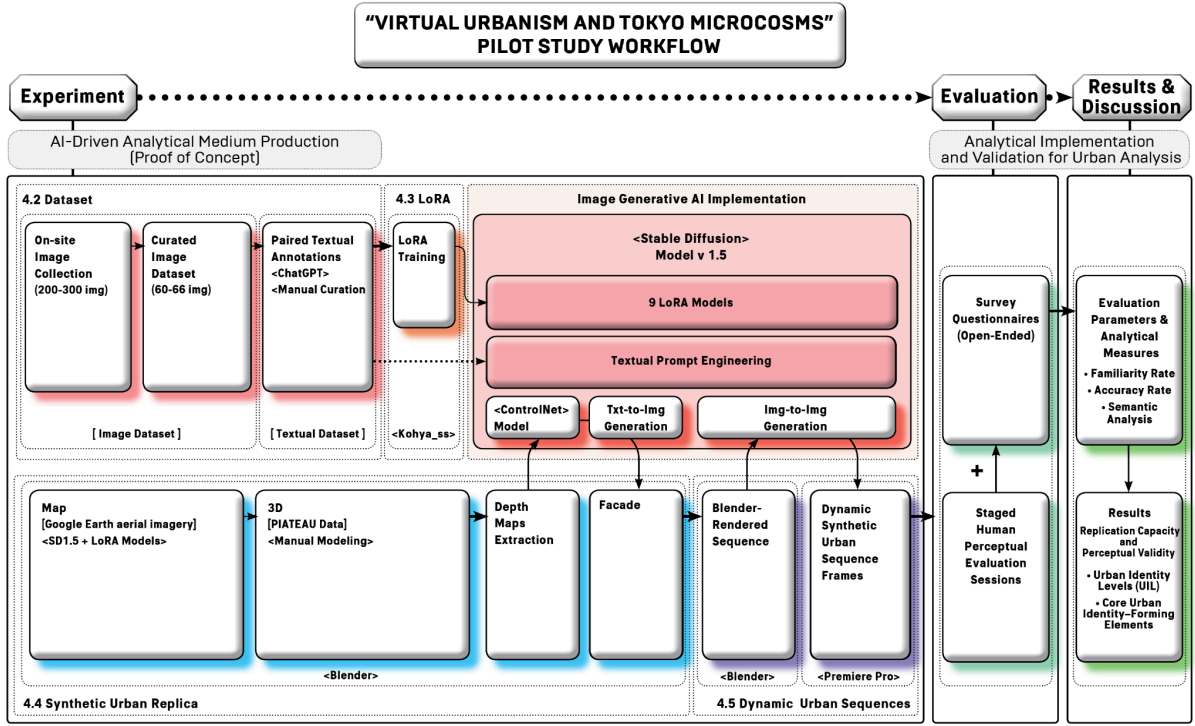

**Figure 1.** Methodological Workflow of the Pilot Study.

## 4. Experiment

The experiment follows a four-stage pipeline. First, LDM is fine-tuned through area-specific dataset creation and LoRA training, resulting in multiple area-specific LoRA adapters. These adapters are then employed to construct a synthetic urban replica and the corresponding dynamic synthetic urban sequences. The full methodological pipeline is described in Sections 4.1–4.5.

### 4.1 Model Pipeline Components

The Experiment is operationalized by leveraging the widely adopted LDM—Stable Diffusion (SD) from StabilityAI, specifically SD version 1.5 (SD 1.5), trained on the LAION-5B dataset [39]. SD is a latent denoising diffusion architecture that predicts and removes noise across timesteps. This study adopts the standard DDPM formulation introduced by Ho et al. [40] and extended to latent space by Rombach et al. [14], without modifying the diffusion objective or noise-prediction mechanism, as model architecture evaluation falls outside the scope of this work.

Proprietary generators (e.g., MidJourney, DALL·E) were excluded in favor of open-source SD 1.5, ensuring reproducibility and architectural flexibility. Although the more recent SDXL model [41] offers higher visual fidelity, it was not adopted, as the production phase of this study predated its release.

Building on the base model, this experiment integrates LoRA [34] to enable targeted, lightweight fine-tuning by introducing a small set of trainable rank-decomposition matrices into the pretrained model, allowing specific stylistic or contextual attributes to be learned while keeping the original model weights frozen, providing efficient specialization with minimal computational cost. ControlNet [36] is further employed to condition generation on structured inputs—such as depth maps, line drawings, or segmentation masks—thereby enabling explicit control over spatial layout and structural coherence.

### 4.2 Dataset Creation

#### 4.2.1 Image dataset

For each of the nine areas, data collection focused on the visible surface appearance—the "external layers" of urban form. In contrast to the common approach in urban-perception studies that rely on automated Street View (SV) image acquisition (e.g., via the SV Static API [42]), this work employed manually collected image datasets obtained through fieldwork. This approach enabled targeted capture of locality-representative environments and ensured coverage of spatial conditions—such as narrow streets and enclosed urban spaces—that are insufficiently represented or absent in SV imagery, yet critical to perceptual identity, particularly in Asian urban contexts.

Across 3–4 field-visits per area, approximately 200–300 photographs were captured, including street-level views, façade exteriors, and selected architectural details. Each dataset was subsequently curated to approximately 60–66 images—a number identified as the minimal, computationally efficient dataset size, enabling LoRA models to reliably capture distinct local urban identity characteristics. Images were captured using an iPhone 12 Pro in both portrait and landscape orientations (4032×3024 px and 3024×4032 px). No resizing was applied, as uniform rectification of images introduces geometric distortions that compromise architectural proportions. In contrast, preliminary tests indicated that preserving native image dimensions improves LoRA training fidelity and enables more accurate variation in scale and proportion during subsequent generation.

Datasets were organized with a consistent typological distribution: approximately 63% street-view images, 35% façade images, and 2% architectural details or interiors. This composition—combining street-views with axonometric façade images—effectively enabled the single LoRA model implementation to generate both façade-focused and full-streetscape outputs in subsequent stages.

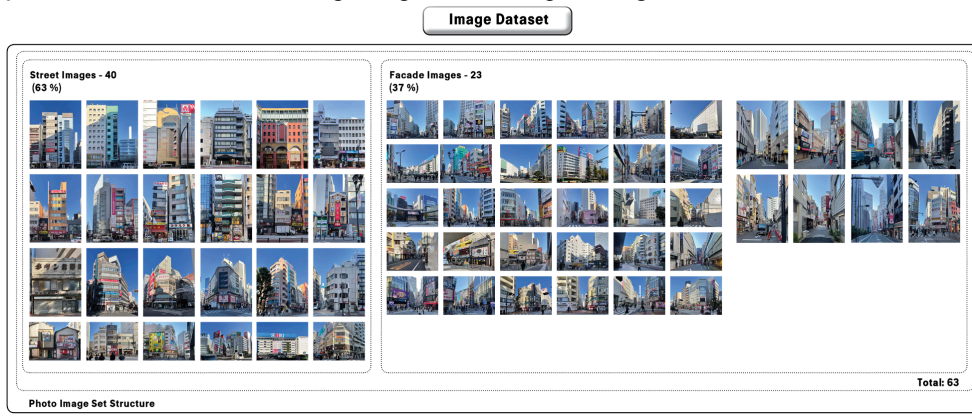

**Figure 2.** Example structure of the image dataset used for training the Ikebukuro.safetensors LoRA model.

### 4.2.2 Textual Dataset

The paired textual dataset was subsequently developed using manually produced captions, refined with ChatGPT assistance, following a structured annotation logic (Figure 3). Although substantially more time-consuming than fully automated methods (e.g., BLIP [43]), manual captioning was adopted as a deliberate methodological decision. The explicit goal of this process was to encode controllable semantic tokens during training, enabling modulation of LoRA influence at inference time via targeted token repetition. This mechanism increases the activation of LoRA-specific features without changing model weights, allowing precise control over aspects of the generated output—such as viewpoint selection and building-scale variation (Figure 4). Importantly, explicit stylistic or interpretive identity labels were not encoded: both the training captions and subsequent prompt engineering were intentionally kept semantically neutral to avoid introducing model bias or prematurely evoking specific architectural styles (e.g., "Gothic"), thereby allowing stylistic and contextual features to emerge through the model's own inferential processes.

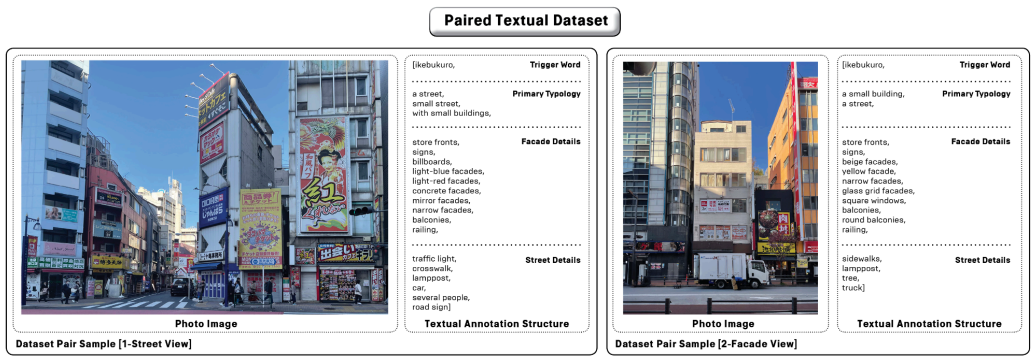

**Figure 3.** Representative image–text pairs demonstrating the structured annotation logic implemented for LoRA training, shown for street-view and axonometric façade image types.

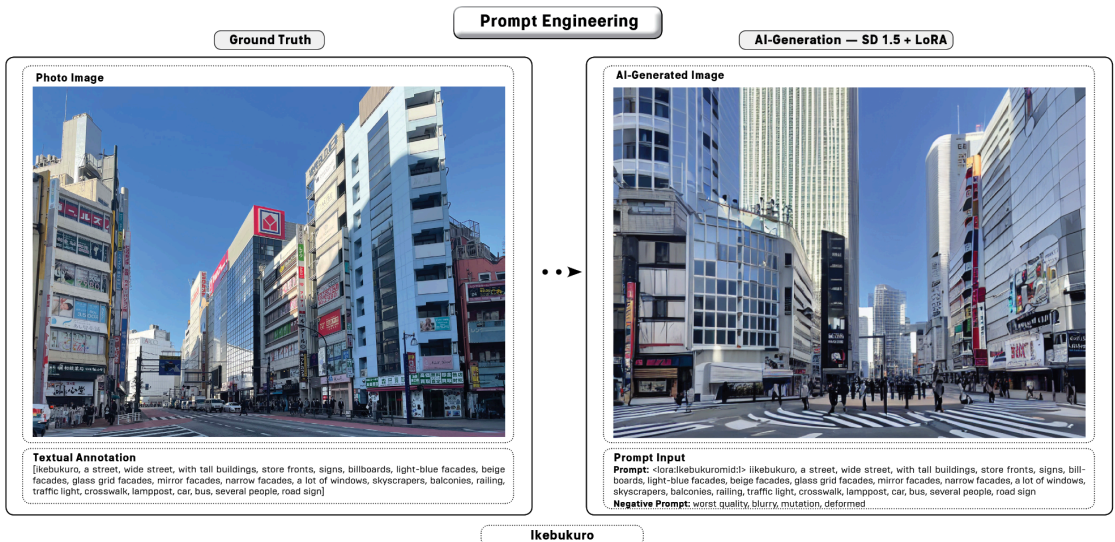

**Figure 4.** Ground-truth image with manual annotation set and a corresponding generative output sample.

### 4.3 LoRA Training Configuration

Following dataset construction, nine LoRA models—one for each of the nine study-areas—were trained locally using the v1-5-pruned.safetensors base model on an NVIDIA GeForce RTX 3080 GPU. The training time for a single model under this configuration was approximately four hours. The v1-5-pruned model was selected due to its optimisation for fine-tuning, efficient VRAM usage, and strong performance in both generalisation and detailed adjustments. As the objective of this study was not to benchmark training configurations—key parameters such as batch size, repeat count, learning rate, and optimizer type were adopted from settings reported in prior studies that demonstrated stable, high-quality results [44]. LoRA training was conducted using Kohya-ss [45], a GUI-based training framework. Most settings remained at their default values, except for the parameters listed below:

• Max Resolution: 768 × 768 pixels
• Epochs: 12
• Batch Size: 2
• Learning Rate: 0.00002

The nine area-specific LoRA models produced in this study are publicly available through an open GitHub repository (see Data Availability section).

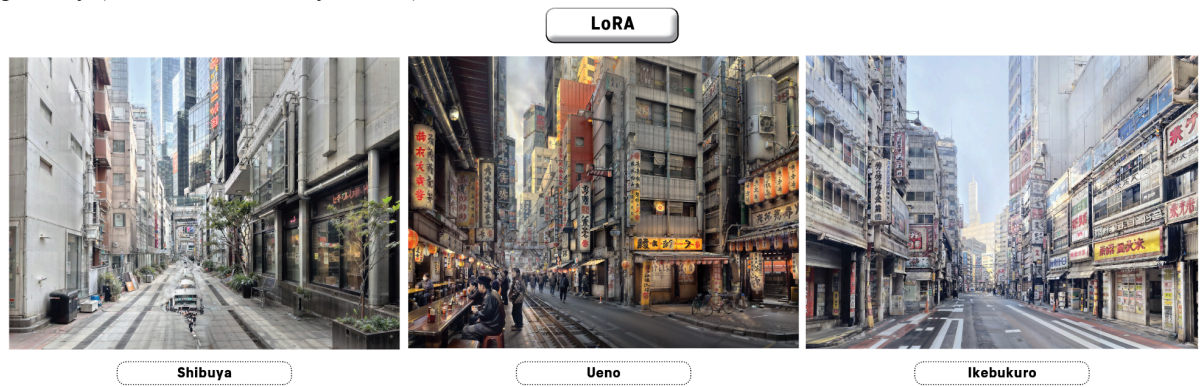

**Figure 5.** Sample images generated during LoRA testing for Shibuya, Ueno, and Ikebukuro areas.

### 4.4 Synthetic Urban Replica Construction

With the area-specific LoRA models established, the experiment advanced to the construction of the synthetic urban replica that underpins subsequent analytical stages. The replica is constructed in three stages, detailed in Sections 4.4.1–4.4.3.

### 4.4.1 Map

The base map was constructed as a controlled abstraction of a 1,500 × 1,500 m segment of Tokyo's urban fabric, incorporating the nine study areas included in the pilot. This approach preserved relative urban density and spatial organization while intentionally simplifying secondary environmental complexity, thereby providing a computationally manageable yet structurally representative base. Consistent with the overall generative pipeline, map construction implemented SD 1.5 and LoRA and was organized into two steps:

**Step 1: Pooled-Area LoRA Fine-Tuning and Base Map Generation**. LoRA training was first conducted using Google Earth aerial imagery pooled from all nine study areas. For each area, 8 aerial samples were extracted and cropped to 1,500 × 1,500 m tiles, producing a total training dataset of 72 images. At this stage, the model learned shared urban characteristics across the nine areas, enabling the generation of synthetic base maps integrating representative features from each zone. Approximately 400 low-resolution map samples were generated, with a single base map selected based on its capacity to represent common environmental characteristics, including relative urban density, scale, and spatial configuration.

**Step 2: Zone-Level LoRA Fine-Tuning and Sector Refinement.** The selected base map was subdivided into 200 × 200 m sectors (64 sectors in total), each assigned to one of the nine study areas based on visual and contextual similarity. A second training phase produced 9 area-specific LoRA models, each trained on 66 Google Earth aerial samples cropped to the sector scale. Each sector was then upscaled and refined using its corresponding area-specific LoRA, resulting in a high-resolution synthetic map with localized spatial characteristics.

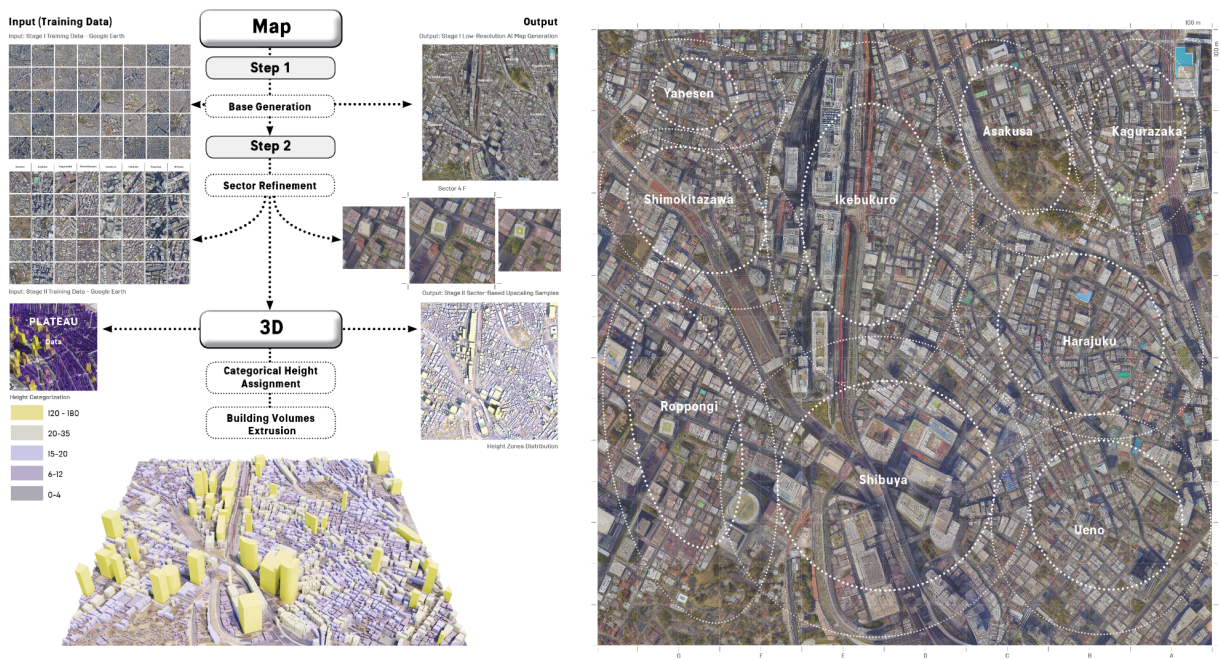

**Figure 6.** Workflow for the production of a 1,500 × 1,500 m synthetic map integrating nine Tokyo case-study areas and its subsequent three-dimensional instantiation.

### 4.4.2 3D
The subsequent stage in synthetic replica development involved transforming the base map from a two-dimensional representation into a three-dimensional model. This was achieved through manual tracing, categorical height assignment, and data import into Blender, where building volumes were extruded using randomized parameters constrained by zone-specific height limits. Height distributions were calibrated using reference data from the PLATEAU platform (MLIT) [46], ensuring contextual fidelity.

### 4.4.3 Façade
In the final stage, façade imagery for the synthetic urban replica was generated using a fine-tuned SD 1.5 model. Sequential camera placement within Blender was used to render depth maps for individual façade segments, which were subsequently imported into SD. Façade generation was performed through image-to-image generation workflow using area-specific LoRA models (see Sections 4.2, 4.3), with ControlNet depth-map based conditioning. Generated façade textures were then re-projected onto the 3D model, resulting in a contextually aligned and visually unified surface representations. (Figure 8)

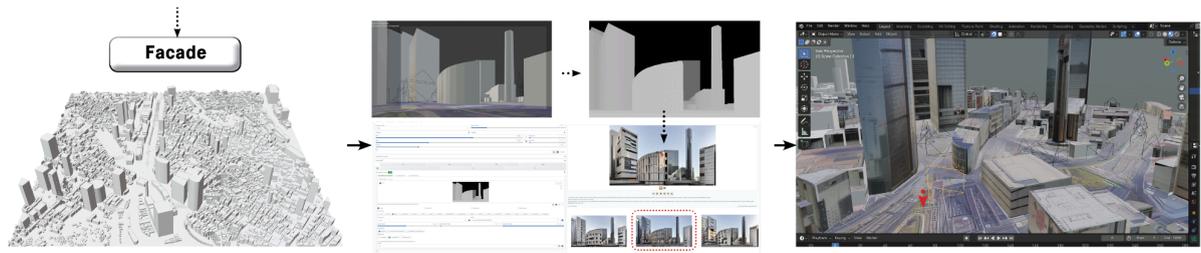

**Figure 7.** Façade generation workflow integrating Blender-based depth extraction and fine-tuned SD 1.5 model with ControlNet conditioning.

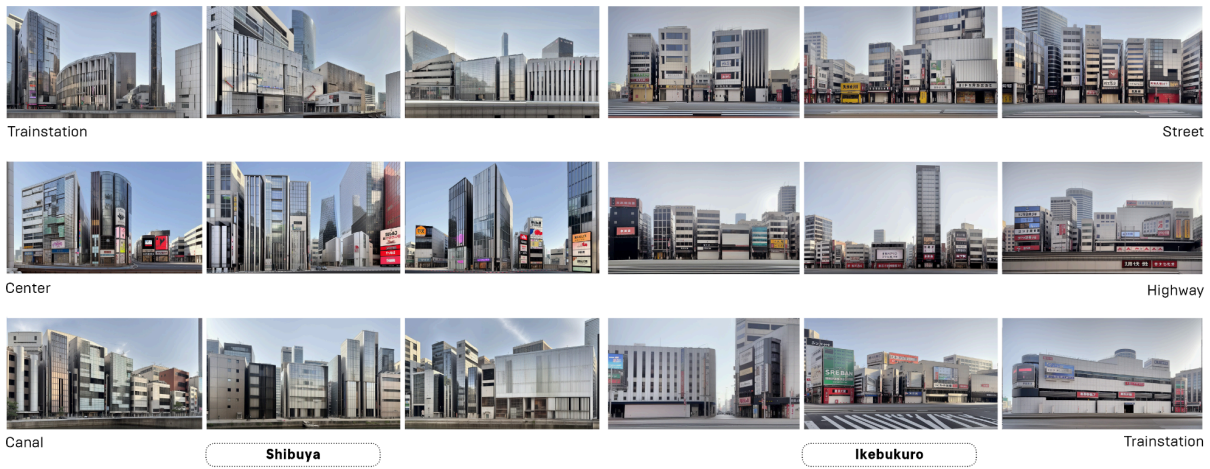

**Figure 8.** Samples of façades generated during synthetic urban replica construction, categorized by case-study area and urban subzone.

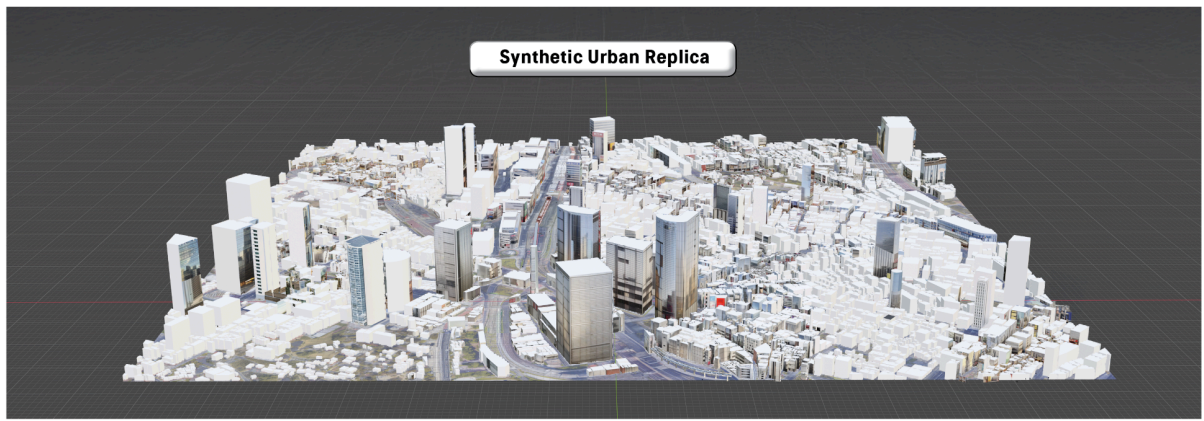

**Figure 9.** Constructed synthetic urban replica: a three-dimensional environment integrating nine Tokyo case-study areas.

### 4.5 Dynamic Synthetic Urban Sequence Production

This section addresses the production of the study's primary analytical medium: dynamic synthetic urban sequences. Within Blender, camera paths were defined inside the synthetic replica environment, with one path corresponding to each case-study area. These paths were rendered as high–frame-rate image sequences and exported for subsequent reprocessing through an image-to-image generation workflow in SD 1.5, implementing the corresponding area-specific LoRA models and prompt configurations. Frame-wise generative transformation was performed under constrained denoising conditions ($\leq 0.68$), ensuring controlled variation within LoRA-defined representational boundaries. Through this process, minor shifts in viewpoint—amplified by the elevated frame rate—were transformed into *successive manifestations* of the same form or environment (Figure 10).

The resulting dynamic synthetic urban sequences constitute the final analytical medium of the pilot study. (Figure 11). Each sequence was standardized as a 30-second video, comprising 385 frames per area (~13 fps), exhibiting the previously defined dynamic quality and landmark-absent conditions. In addition to their analytical function, the sequences form continuous urban-narrative traversals, subsequently used as stimulus material for human perceptual evaluation and public showcase sessions. The complete set of dynamic synthetic urban sequences is publicly available for viewing; access details are provided in the Data Availability section.

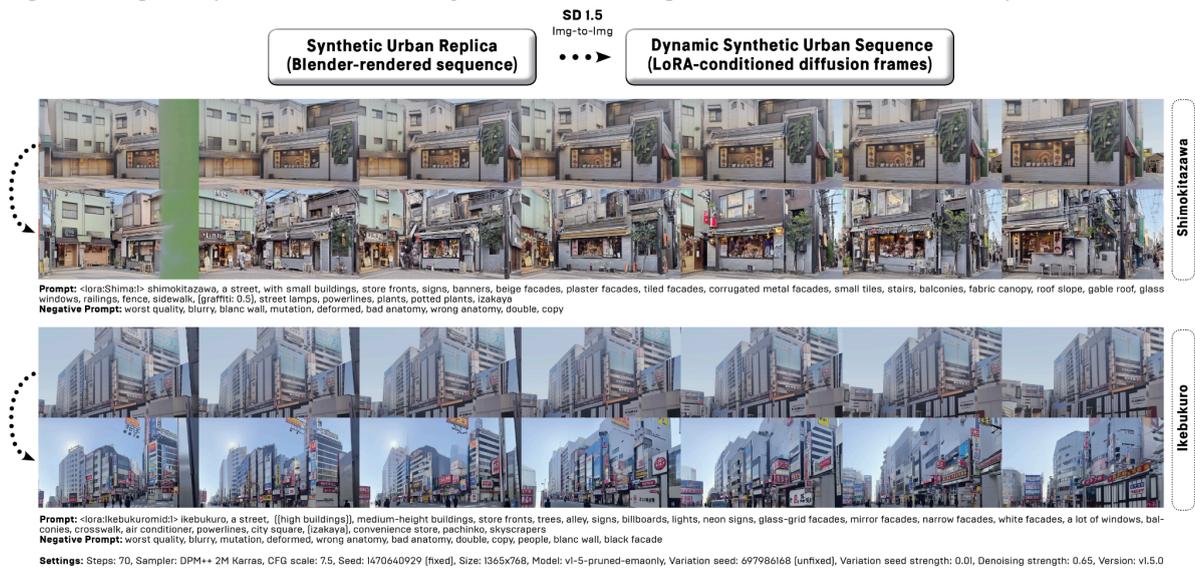

**Figure 10.** Image-to-image diffusion–based (SD 1.5, LoRA) transformation of a Blender-rendered urban replica footage to dynamic synthetic urban sequence, illustrated through a subset of frame samples from two study areas.

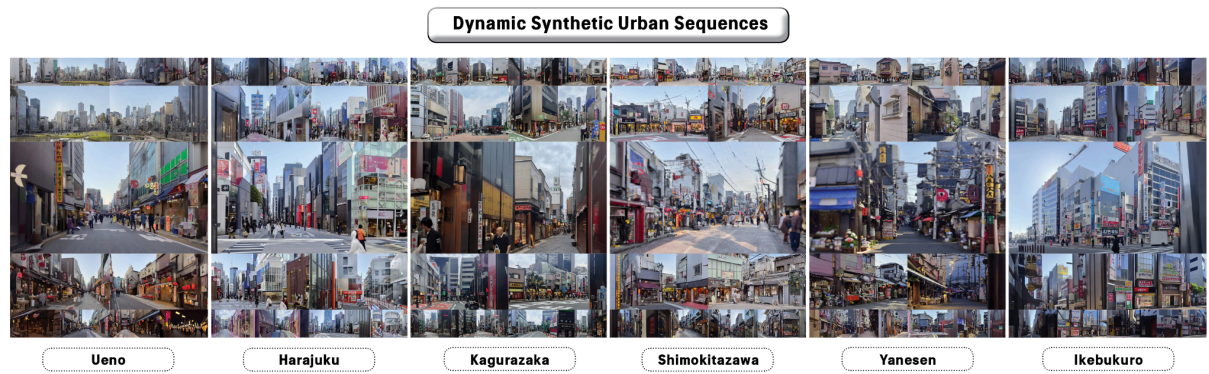

**Figure 11.** Representative frames from the dynamic synthetic urban sequences.

## 5. Evaluation

This section outlines the methodology for perceptual evaluation conducted using the materials produced in Section 4, detailing the evaluation design and the introduced analytical parameters.

Unlike conventional survey-based perceptual studies that rely on static stimuli and predefined evaluative descriptors, the present work leveraged AI-driven dynamic synthetic urban sequences to frame the research survey as an immersive experience, with experimental materials functioning as a front-end communicative medium that simultaneously supported data collection, interpretation, and dissemination (Figure 12). This design enabled open-ended, cue-emergent response elicitation, allowing urban identity to be examined through participant perception while effectively avoiding imposed taxonomies.

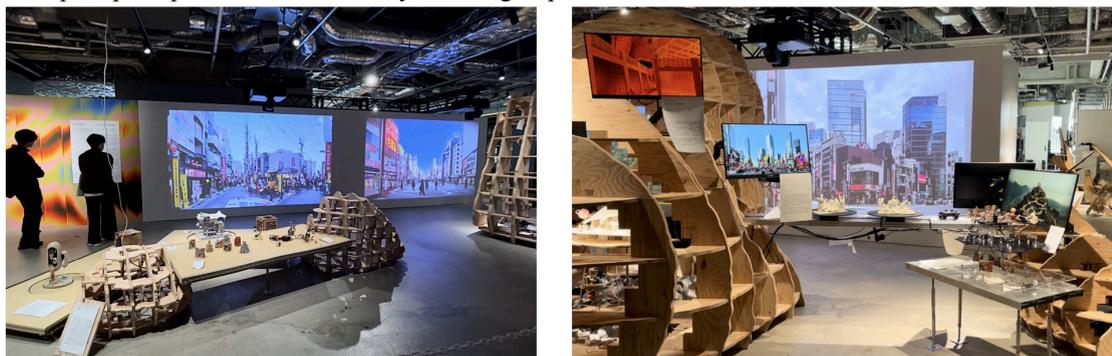

**Figure 12.** Visual material produced within the pilot study, Tokyo Creative Forum 2025.

## 5.1 Method
### 5.1.1 Evaluation Procedure
The evaluation procedure was designed as a staged human perceptual assessment. Participants were presented with dynamic synthetic urban sequence videos, accompanied by surveys subsequently analysed using both quantitative and semantic methods. The survey framework was informed by *Situationist International*'s *Questionnaire for Readers* [49] and Kevin Lynch's *The Image of the City* [50], and was structured to address the previously stated study objectives. The evaluation process comprised three phases: pre-viewing familiarity profiling (≈5 min); initial familiarisation (≈15 min), involving two looped viewings of all nine sequences; and in-depth analysis (≈1 h), involving five looped viewings per sequence during questionnaire completion.

**Table 1.** Condensed survey instrument and analytical role of questionnaire items

| No. | Analytical Role | Question |
|---|---|---|
| 0 | Familiarity Rate Parameter | Please indicate your familiarity with each of the areas listed below by underlining the option that best applies to you. (Not Familiar, Quick Visits, Regular Attendance, and Continuous Residence) |
| 1 | Accuracy Rate Parameters | What district do you think the current film is depicting? |
| 2 | Semantic Analysis | What made you think it was this particular district, what made it feel familiar to you? |
| 3 | Semantic Analysis | Are there any specific visual elements or features that led you to this conclusion? |
| 4 | Semantic Analysis | Are there any elements that feel 'wrong' or out of place? |
| 5 | Semantic Analysis | What features could be adjusted or added to make the location more recognizable in the video? |

### 5.1.2 Evaluation Parameters and Analytical Measures
Responses were structured to enable comparative analysis across sequences and participant groups, organized by the nine dynamic synthetic urban sequences (corresponding to the nine Tokyo study areas) and segmented into three participant groups: General (entire cohort), Local (Japanese nationals), and Foreign (foreign residents of Tokyo). For the analysis presented in the subsequent Section 6, a set of evaluation parameters is introduced.

**Familiarity Rate Parameter**

Pre-viewing familiarity profiling captured participants' familiarity with each of the nine study areas, assessed using a four-level scale—Not Familiar, Quick Visits, Regular Attendance, and Continuous Residence—which was converted into a continuous Familiarity Rate Parameter using normalized exposure weights (0.0, 0.4, 0.7, and 1.0, respectively). The Familiarity Rate was calculated as the mean weighted familiarity score across respondents and reported as a percentage (Equation 1).

$$FR_{a,g} = \frac{1}{N_{a,g}} \sum_{i=1}^{N_{a,g}} w(F_{i,a}) \qquad (1)$$

Where: $FR_{a,g}$ = Familiarity Rate for area $a$ and participant group $g$; $N_{a,g}$ = number of participants in group ; $F_{i,a}$ = familiarity level selected by participant $i$ for area $a$; $w(\cdot)$ = familiarity weight mapping $\{0, 0.4, 0.7, 1.0\}$

This parameter was introduced to contextualize identification performance in relation to participant prior familiarity, in conjunction with the Accuracy Rate. (Figure 14)

**Accuracy Rate Parameters**

Following sequence viewing, participants assigned each of the nine showcased dynamic synthetic urban sequences to one of the nine Tokyo study areas. Identification performance was quantified as an Accuracy Rate, defined as the proportion of correct assignments expressed as a percentage (Equation 2).

$$AR = \frac{C}{T} \times 100 \qquad (2)$$

Where: $C$ = number of correct sequence–area assignments; $T$ = total number of assignments considered

- **Accuracy Rate Parameter per Participant**

The Accuracy Rate calculated per participant was introduced to evaluate the overall perceptual validity of the produced synthetic urban sequences, assessing the extent to which participants were able to correctly associate the sequences with their corresponding real-world areas, thereby operationalizing study Objectives (a) and (b).

- **Accuracy Rate Parameter per Sequence = Urban Identity Level (UIL)**

The Accuracy Rate calculated per sequence was introduced to measure how consistently each dynamic synthetic urban sequence was identifiable as its corresponding real-world area across all participants. As each sequence represents a specific study area, this parameter additionally functions as a preliminary Urban Identity Level (UIL) metric, operationalizing the relative visual distinctiveness of each area within the study set addressing Objective (cI).

**Semantic Analysis**

The final analytical stage employed a qualitative semantic analysis of participants' free-text survey responses, operationalized by frequency-based assessment of recurrent terms and phrases associated with correct perceptual recognition. The extracted lexical patterns were subsequently quantified and organized into five thematic groups—Elements, Environment, Typology, Color, and Qualitative Characteristics—defining the key recognition catalysts, previously defined as core urban identity–forming elements, for each analyzed Tokyo area (Table 2)—thereby operationalizing Objective (cII).

## 6. Results and Discussion

This section reports and interprets the data acquired during the pilot study evaluation sessions to assess the perceptual legitimacy of the synthetic urban replicas—here operationalized as dynamic synthetic urban sequences—and their applicability as an analytical medium for examining urban identity.

### 6.1 Participants characterization

A total of 36 participants (P1-P36) took part in the evaluation stage across three film-viewing sessions. Participants ranged from 21 to 42 years old and were primarily architecture-affiliated, including students, architects, and professionals from engineering, academia, and general employment. AI familiarity varied, with most reporting occasional exposure and a smaller subset indicating regular academic or professional use. The sample comprised 20 Japanese participants (56%) and 16 foreign residents (44%). Length of residence in Tokyo ranged from less than one year to over five years (≤1 year: 3 participants ; 1–3 years: 14 ; 3–5 years: 5; ≥ 5 years: 14).

### 6.2 Generative Replication Capacity and Perceptual Validity of Synthetic Replicas

Across participants, Accuracy Rates per participant ranged from 44% (3 participants) to 100% (14 participants). Where the majority of participants demonstrated high identification performance, with a mean Accuracy Rate of approximately 81% (Figure 13).

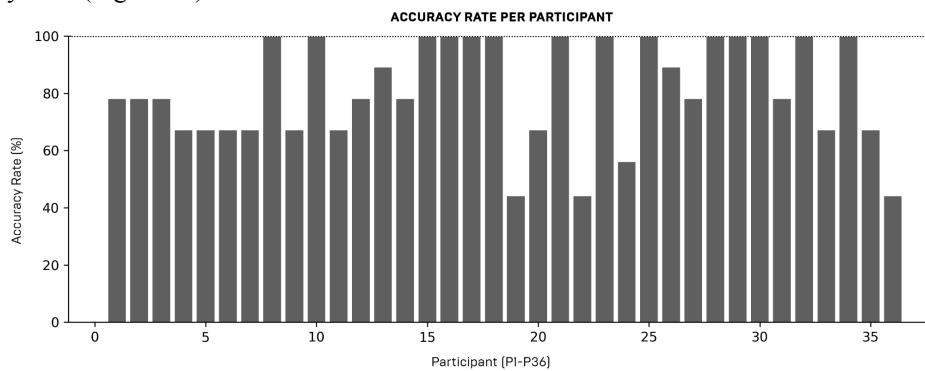

**Figure 13.** Accuracy Rate per participant, showing individual variation in correct sequence–area identification across the participants (P1-P36).

These results indicate that the synthetic environments were, on average, highly identifiable. Within the context of this pilot study, this provides strong preliminary evidence of the capacity of generative models to replicate real urban settings in a manner that supports consistent perceptual recognition.

## 6.3 Analytical Applicability of Synthetic Urban Replicas
## 6.3.1 Accuracy-Based Evaluation of Preliminary Urban Identity Levels (UIL)

This section assesses the applicability of AI-driven synthetic urban replicas as an analytical medium for studying urban identity.

Accuracy Rate parameters calculated per sequence—operationalized in this study as the Urban Identity Level (UIL) metric—were compared across the three participant groups (General, Local, and Foreign) and used to rank the study areas according to relative identifiability, from higher to lower UILs. These rankings were then examined alongside participants' Familiarity Rates to verify that identifiability was not solely attributable to prior exposure (Figure 14).

| GENERAL | | | LOCAL PARTICIPANTS | | | FOREIGN PARTICIPANTS | | |
|---|---|---|---|---|---|---|---|---|
| Accuracy Rate (UIL) [%] | Familiarity Rate [%] | | Accuracy Rate (UIL) [%] | Familiarity Rate [%] | | Accuracy Rate (UIL) [%] | Familiarity Rate [%] | |
| **Asakusa** 100 ▲ | Shibuya 69 ▼ | | **Asakusa** 100 ▲ | Shibuya 69 ▼ | | Asakusa 100 | Shibuya 68 | |
| Harajuku 100 | Ueno 62 | | **Harajuku** 100 ▲ | Ueno 63 | | **Harajuku** 100 ◀‥▶ | **Harajuku** 63 | |
| **Shimokitazawa** 86 ▲ | Harajuku 60 | | Ueno 90 | Shimokitazawa 61 | | Shimokitazawa 81 | Ikebukuro 62 | |
| **Shibuya** 83 ▼ | Ikebukuro 58 | | Shimokitazawa 90 | **Harajuku** 57 ▲ | | Shibuya 75 | Ueno 60 | |
| Ikebukuro 75 | **Shimokitazawa** 53 ▲ | | **Shibuya** 90 ▼ | **Asakusa** 57 ▲ | | **Roppongi** 63 ◀‥▶ | **Roppongi** 46 | |
| Ueno 75 | **Asakusa** 51 ▲ | | Ikebukuro 90 | Ikebukuro 54 | | Ueno 56 | Shimokitazawa 44 | |
| Roppongi 72 | Roppongi 49 | | Kagurazaka 85 | **Roppongi** 51 ▼ | | Ikebukuro 56 | Asakusa 44 | |
| Yanesen 69 | Kagurazaka 32 | | Yanesen 85 | Kagurazaka 44 | | **Yanesen** 50 ◀‥▶ | Kagurazaka 19 | |
| Kagurazaka 67 | Yanesen 19 | | **Roppongi** 80 ▼ | Yanesen 32 | | **Kagurazaka** 44 | Yanesen 5 | |

**Figure 14.** Comparative ordering of Accuracy Rate (UIL) and Familiarity Rate across dynamic synthetic urban sequences (study areas), shown separately for the General, Local, and Foreign participant groups.

**(Figure 14) Note 1.** Boldface highlights the study areas discussed in detail in the subsequent analysis. Directional arrows indicate the relationship between Accuracy Rate (UIL) and Familiarity Rate, with arrow direction denoting relative alignment or divergence.

**(Figure 14) Note 2.** Developmental-origin ordering of study areas from most organic to most corporate (included to support interpretation of the following analysis): Shimokitazawa, Harajuku, Yanesen, Kagurazaka, Asakusa, Ueno, Shibuya, Ikebukuro, Roppongi.

Assessment of the results shown in Figure 14 revealed a clear relationship between the developmental origins of the study areas and their corresponding UIL. Organically evolved districts consistently achieved the highest UIL—such as Harajuku (100%) and Shimokitazawa (86%)—whereas areas shaped primarily by corporate redevelopment exhibited notably lower scores, including Ikebukuro (75%) and Roppongi (72%) (values reported for the General group).

This pattern becomes more pronounced when considered alongside Familiarity Rates. While the Foreign participant group shows a clear dependency between Familiarity and Accuracy Rate (UIL), analysis of the Local and aggregated General data reveals patterns that cannot be explained by familiarity alone. Notably, Shibuya—despite being the most familiar area—ranks comparatively lower in UIL, a pattern similarly observed for Roppongi. Conversely, organically developed areas such as Shimokitazawa and Harajuku exhibit lower Familiarity Rankings yet consistently higher Accuracy Rates across all participant groups, effectively demonstrating elevated UIL values.

The results indicate that, at a deeper perceptual level, corporately developed areas exhibit comparatively lower UIL: despite higher participant familiarity, these areas are less consistently identifiable when viewed as orientation marker–absent synthetic replicas. In contrast, organically developed areas display visibly stronger UIL values, reflecting more robust contextual locality and authenticity that are evidenced independently of singular landmarks.

Importantly, while the observed relationship between developmental origin and urban identity is not in itself a novel theoretical conclusion, the primary contribution of this study lies in the ability to measure, assess, and cross-reference such differences through an explicitly defined quantitative metric. The introduction of Urban Identity Level (UIL) establishes a novel operational metric through which urban identity can be treated as a quantifiable analytical construct within computational analysis.

## 6.3.2 Semantic Identification of Core Urban Identity–Forming Elements

This section examines the semantic basis underlying correct identification of synthetic urban sequences as their corresponding real-world areas, based on analysis of recurrent descriptors, spatial features, and visual impressions in participants' free-text responses (Table 1, Q2–Q5). The resulting core urban identity-forming elements, organized by category, and citation frequencies are summarized in Table 2.

| Shimokitazawa | Typology - Shops [31] | Element - Clothing [25] | Quality - Small [17] |
|---|---|---|---|
| Harajuku | Typology - Shops [24] | Quality - Colorful [22] | Element - Fashion [12] |
| Yanesen | Quality - Old [16] | Typology - Private House [12] | Characteristic - Narrow [10] |
| Kagurazaka | Quality - Traditional [12] | Element - River [10] | Characteristic - Narrow [10] |
| Asakusa | Color - Red [38] | Quality - Traditional [23] | Typology - Temples [10] |
| Ueno | Environment - Park [21] | Typology - Izakaya [13] | Characteristic - Wide [14] |
| Shibuya | Element - Signage [32] | Characteristic - High [25] | Environment - River [20] |
| Ikebukuro | Element - Signage [21] | Quality - Cluttered [10] | Color - Red [9] |
| Roppongi | Material - Glass [10] | Typology - "Glass" Buildings [10] | Element - Bridge [10] |

**Table 2.** Core urban identity–forming elements by area and citation frequencies derived from semantic analysis.

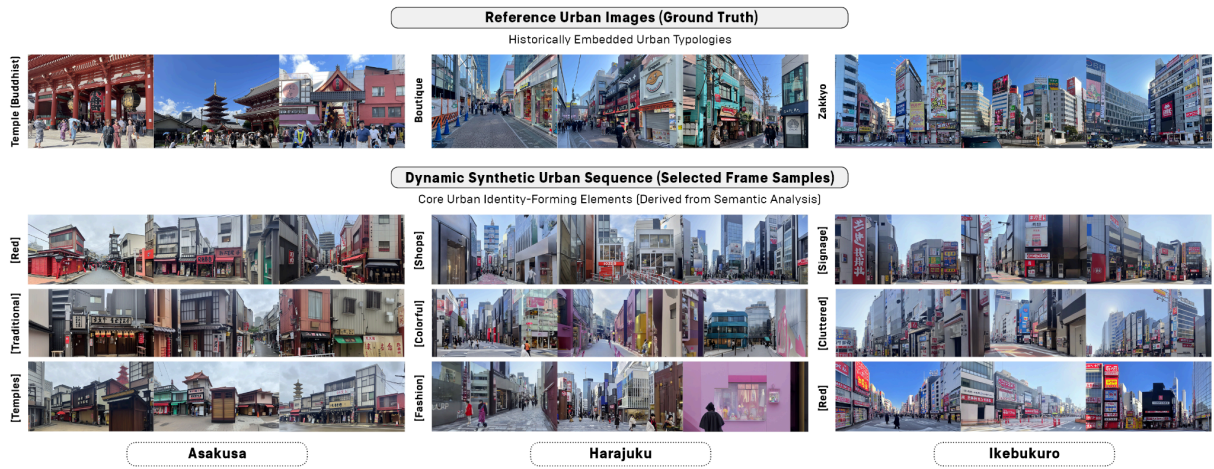

**Figure 15.** Visual corroboration of semantically identified core urban identity–forming elements with historically embedded urban typologies through dynamic synthetic urban sequence frame samples.

Semantic analysis revealed that the most consistently cited recognition catalysts—defined in this study as core urban identity–forming elements—collectively correspond to typologies that are culturally and historically embedded in each area's context (for example, Asakusa [red] + [traditional] + [temples], i.e., Buddhist Temple architecture). Comparable typological patterns were identified across multiple study areas—zakkyo-style mixed-use buildings in Ikebukuro, small private establishments in Kagurazaka, local groceries in Yanesen, izakaya clusters in Ueno, and fashion-oriented retail in Harajuku. (Figure 15)

Importantly, these elements—and the corresponding typological identifications—emerged within the dynamic synthetic urban sequences, despite typological fidelity not being explicitly imposed or enforced. Instead, typological coherence resurfaced through feature combinations that the LDM internalized during LoRA training and reproduced within the sequences with sufficient repetition density to become perceptually identifiable—*essence revealed through successive manifestations* [7].

As in the previous section, although the identified associations may initially appear self-evident, their significance lies precisely in their established recognizability. This correspondence between derived results and well-documented urban characteristics positions the pilot study as a proof of concept, confirming the methodological validity of the proposed approach. When operationalized through AI-driven synthetic replicas endowed with dynamic qualities—Novak's interrepresentational approach is reintroduced as a viable AI-driven analytical method for identifying core urban identity–forming elements, constituting a foundational analytical component of the VU framework. Together, these findings provide empirical grounding for the VU framework as a novel method of urban analysis.

### 7. Limitations and Challenges

The pilot study revealed several methodological and technical limitations that delineate clear directions for future refinement and extension of the VU framework.

Manual, non-automated workflow: The construction of synthetic urban replicas and dynamic sequences relied on extensive manual curation across all stages of the pipeline. While sufficient for a pilot implementation, this approach limits scalability and methodological consistency. Future iterations will require automated data acquisition, standardized dataset construction, and reproducible validation protocols.

Single-modality AI implementation and limited identity parameterization: The pilot pipeline relied primarily on image-generative latent diffusion models, with large language models serving only auxiliary roles, thereby constraining urban identity assessment to visual perception and surface-level morphological cues. Future iterations should adopt a modular, multimodal AI architecture integrating linguistic, visual–semantic, and spatially grounded generative systems, together with expanded fine-tuning datasets and advanced generative model architectures, to support multi-parameter urban identity metrics encompassing perceptual, semantic, narrative, and behavioral dimensions.

Non-automated, open-loop evaluation structure: While the evaluation design successfully prioritized participant engagement and open-ended perceptual interpretation, it relied on a relatively small cohort and in-person, non-automated data collection. Future work will require scalable, platform-based evaluation frameworks supported by automated semantic analysis, as well as closed-loop workflows in which evaluation outputs are reintegrated into the generative pipeline, enabling iterative refinement of both representations and urban identity metrics.

## 8. Conclusion

The pilot study successfully addressed all three predefined objectives (a)–(c), as demonstrated in Sections 5 and 6. Specifically, the results confirmed the technical feasibility of AI-driven replication of urban environments, perceptual validity of the resulting synthetic urban replicas and their applicability as an analytical medium, thereby establishing an empirical proof of concept for the Virtual Urbanism (VU) framework.

As previously stated, a central actionable aim of the VU framework is the development of computationally tractable and analytically operational urban identity metrics. This pilot study operationalized this aim through the introduction of the Urban Identity Level (UIL) (see Section 6.3.1), presented as a preliminary metric that renders urban identity cross-comparable and analytically operational. Building on this foundation, further development of the VU framework has the potential to advance computational urban perception research by integrating qualitative, perceptual dimensions of the urban environment—such as local identity and perceived authenticity—into measurable analytical constructs suitable for expansion into automated, multi-parameter metrics and reintegration into AI-assisted urban research and generative urban design (GUD) workflows.

Beyond its empirical contributions, the study served to articulate the conceptual positioning of artificial intelligence within the VU framework. This was demonstrated through the use of generative AI as an instrument supporting the production of a novel analytical medium for urban identity inquiry—one explicitly designed to enable forms of insight not directly attainable through engagement with the real urban environment itself (see Sections 1.1 and 6.3.2). The AI-augmented methodological core of the VU framework is thus positioned as the iterative, pipeline-based development of a modular analytical system, enabling the progressive integration of multimodal AI tools for translating high-dimensional, large-volume, and weakly structured urban data into analytical representations accessible to human cognition. Within this framework, AI functions as an intermediary layer, mediating between large-scale urban data and human perceptual and interpretive capacities, and providing a foundation for the continued evolution of urban identity-focused computational analytics within the broader field of architectural computing.

## Data Availability

The visual materials produced within the pilot study are publicly available for viewing:
via Vimeo (https://vimeo.com/1053283654/677f8ab5f3)
and YouTube (https://www.youtube.com/watch?v=s4ENIZPcUKw).
The trained Stable Diffusion v1.5–compatible LoRA models corresponding to the nine Tokyo study areas, which constitute the primary dataset generated in this study, are available for download through an open GitHub repository (https://github.com/glinskayamaria/Virtual-Urbanism-Library).